\documentclass[letterpaper, 10 pt, journal, twoside]{ieeetran}  

\IEEEoverridecommandlockouts                              

\usepackage[utf8x]{inputenc}
\usepackage{amsmath}
\usepackage{amssymb}
\usepackage{wrapfig}
\usepackage{graphicx}
\usepackage{algorithm}
\usepackage{algpseudocode}
\usepackage{eqparbox}
\usepackage{mdwmath}
\usepackage{mdwtab}
\usepackage{array}
\usepackage{authblk}
\usepackage{bm}

\usepackage{hyperref}
\usepackage{units}
\usepackage{multirow}

\usepackage{subfigure}

\usepackage{setspace}

\usepackage[left=2cm,right=2cm,top=2cm,bottom=2cm]{geometry}

\usepackage[colorinlistoftodos, textwidth=3.7cm]{todonotes}
\definecolor{MyBrickRed}{cmyk}{0,0.89,0.94,0.28}

\usepackage[english]{babel}

\usepackage{pslatex}

\graphicspath{
	{figs/}
    {evaluation/}
	{matlab/}
	{anim/}
}

\renewcommand{\vec}[1]{\mathbf{#1}}

\renewcommand\IEEEkeywordsname{Keywords}

\title{\LARGE 
CyberCortex.AI: An AI-based Operating System for Autonomous Robotics and Complex Automation
}

\author{
Sorin~Grigorescu$^{1}$ and Mihai~Zaha$^{1}$
\thanks{
Manuscript received: February 22, 2024.}
\thanks{
$^{1}$ The authors are with \href{https://www.cybercortex.ai/}{CyberCortex Robotics} and the \href{https://www.rovislab.com/}{Robotics, Vision and Control Laboratory (RovisLab)}, Transilvania University of Brasov, Romania. Corresponding author email: s.grigorescu@unitbv.ro}
\thanks{
Digital Object Identifier (DOI): see top of this page.}
}

\begin{document}

\maketitle

\begin{abstract}

The underlying framework for controlling autonomous robots and complex automation applications are Operating Systems (OS) capable of scheduling perception-and-control tasks, as well as providing real-time data communication to other robotic peers and remote cloud computers. In this paper, we introduce CyberCortex.AI, a robotics OS designed to enable heterogeneous AI-based robotics and complex automation applications. CyberCortex.AI is a decentralized distributed OS which enables robots to talk to each other, as well as to High Performance Computers (HPC) in the cloud. Sensory and control data from the robots is streamed towards HPC systems with the purpose of training AI algorithms, which are afterwards deployed on the robots. Each functionality of a robot (e.g. sensory data acquisition, path planning, motion control, etc.) is executed within a so-called DataBlock of Filters shared through the internet, where each filter is computed either locally on the robot itself, or remotely on a different robotic system. The data is stored and accessed via a so-called \textit{Temporal Addressable Memory} (TAM), which acts as a gateway between each filter's input and output. CyberCortex.AI has two main components: \textit{i}) the CyberCortex.AI.inference system, which is a real-time implementation of the DataBlock running on the robots' embedded hardware, and \textit{ii}) the CyberCortex.AI.dojo, which runs on an HPC computer in the cloud, and it is used to design, train and deploy AI algorithms. We present a quantitative and qualitative performance analysis of the proposed approach using two collaborative robotics applications: \textit{i}) a forest fires prevention system based on an Unitree A1 legged robot and an Anafi Parrot 4K drone, as well as \textit{ii}) an autonomous driving system which uses CyberCortex.AI for collaborative perception and motion control.

\end{abstract}



\section{Introduction}
\label{sec:introduction}

\IEEEPARstart{D}{espite} many developments in Artificial Intelligence (AI) and software frameworks, the Operating Systems (OS) currently used in robotics have not changed over the last 15 years~\cite{Kramer_2007}. These are typically overengineered and lack the encapsulation of high-bandwidth data streaming techniques required for the design, training and continuous deployment of the AI algorithms used within the robots' perception-and-control pipelines. Real-time communication and data management is particularly important in the age of Artificial Intelligence (AI), since the development and evaluation of AI algorithms is dependent on high quantities of sensory and control information.

Over the last decade, AI and Deep Learning~\cite{LeCun_2015} became the main technologies behind many novel robotics applications, ranging from autonomous navigation and self-driving cars~\cite{Grigorescu_2020} to dexterous object manipulation~\cite{OpenAI_2018} and natural language processing~\cite{brown2020language}. The key ingredient in a deep learning algorithm is the data collected from the robots themselves, data which is required for training the Deep Neural Networks (DNN) running within the perception-and-control pipelines of the robots.

\begin{figure}
    \centering
	\begin{center}
		\includegraphics[scale=1.0]{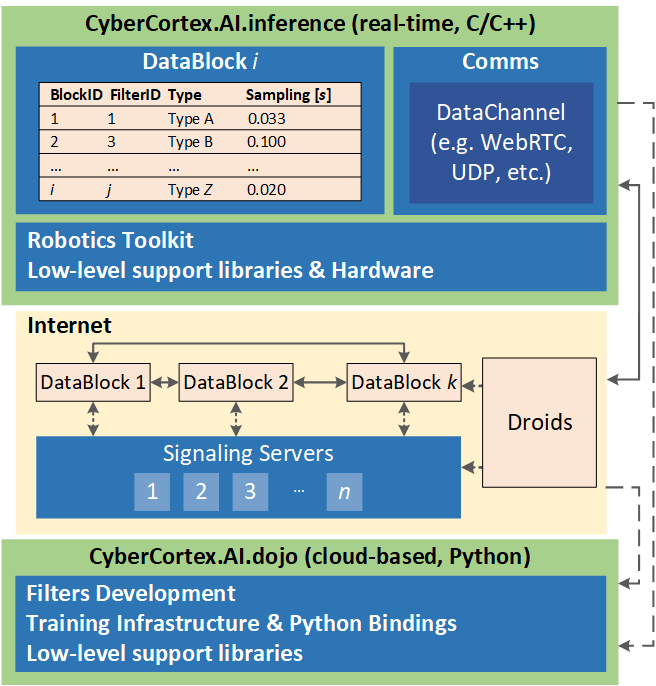}
        \vspace{-1mm}
        \caption{\textbf{The CyberCortex.AI operating system software stack}, composed of \textit{i}) a scalable inference platform executing a DataBlock of so-called Filters and \textit{ii}) a dojo used for designing, training and deploying the Filters and their underling AI algorithms. The Filters from different DataBlocks exchange information via the internet by establishing peer-to-peer communication using a DataChannel and a set of redundant Signaling Servers. The dotted lines are used during development and training.}
        \label{fig:block_diagram_simple}
	\end{center}
    \vspace{-8mm}
\end{figure}

Even with the latest decade's advances in AI, current robotic operating systems still lack a data-driven architecture that would enable rapid AI-based application development for robotics. ROS (Robotic Operating System)~\cite{Quigley_ROS_2009}, the current de-facto middleware for building robotic software, provides many classical (non-learning) algorithms in the form of packages, but lacks the architectural design which would enable the rapid development, retraining and testing of AI technologies for robotics. To meet these challenges, many developers have created a wide variety of AI software stacks that are either used alongside ROS, or are replacing ROS altogether. Such an example is the binding between ROS and Microsoft's AI platform~\cite{MicrosoftAI_2018}, which are two different solutions designed with different philosophies in mind.

In this paper, we introduce the CyberCortex.AI robotics operating system from Fig.~\ref{fig:block_diagram_simple}. It is composed of two platforms: \textit{i}) the \textit{CyberCortex.AI.inference} system, located on the robots' embedded computing devices, and \textit{ii}) the \textit{CyberCortex.AI.dojo} system, executed on High Performance Computers (HPC) used for training the AI algorithms. The CyberCortex.AI.inference system on each robot is a real-time implementation of a so-called \textit{DataBlock}. A DataBlock is a table-like structure identified through an unique ID, where each entry is a so-called \textit{Filter}.

A filter delivers the result of a specific algorithm (e.g. data acquisition, perception, sensor fusion, motion planning, etc.) at specified sampling time rates, while a robotic application is implemented by interconnecting different filters. The outputs of the filters are either computed locally on the robot, or are received from other robots through a DataChannel communication pipeline. The BlockID column of the DataBlock illustrated in Fig.\ref{fig:block_diagram_simple} indicates the DataBlock ID of the robot calculating the outputs of the filters on the local DataBlock \textit{i}.

The DataChannel communication pipeline is also implemented as a filter, allowing for interchangeable data transfer protocols. In our work, we have routed the communication through the internet using the WebRTC protocol~\cite{Sredojev_2015}. The WebRTC implementation of the DataChannel enables direct peer-to-peer communication, which is established via a set of redundant Signaling Servers that are facilitating the discovery of robots between themselves, once a robot starts its DataBlock and registers it in the signaling servers. For training the AI algorithms, we stream the data from the DataBlocks to the CyberCortex.AI.dojo, which offers the software infrastructure required to design, train, deploy and maintain the AI modules. The dotted lines in Fig.~\ref{fig:block_diagram_simple} represent communication channels which are solely used for acquiring training data and updating the AI modules running in the DataBlocks.

The design principles of CyberCortex.AI are:

\begin{itemize}
    \item high-bandwidth data streaming between heterogeneous robotic systems, as well as between robots and the High Performance Computers (HPC) used for training AI algorithms;
    \item data storage and access using a so-called \textit{Temporal Addressable Memory} (TAM);
    \item a protocol agnostic data communication system;
    \item cross-platform and scalability; CyberCortex.AI is currently supported for the Windows, Linux and Android operating systems, while running on embedded computing devices like Nvidia Jetson AGX Xavier and Raspberry PI, as well as on high-performance computers;
    \item upgradable software infrastructure for training and deployment of AI algorithms;
    \item automatic discovery of robot peers and establishment of direct peer-to-peer communication;
    \item data processing location abstractization;
    \item self-installation of software modules;
    \item DataBlock management via threading, as opposed to inter-process communication;
    \item rapid prototyping via internet communication tools and debugging.
\end{itemize}

The rest of the paper is organized as follows. Section~\ref{sec:related_work} covers the related work, while the architecture of CyberCortex.AI is presented in Section~\ref{sec:architecture}. Performance evaluation for two experimental use-cases is given in Section~\ref{sec:evaluation}. Finally, conclusions are stated in Section~\ref{sec:conclusions}.

\section{Related Work}
\label{sec:related_work}

In the following, we provide an overview of software frameworks for robotics, dedicating special chapters on ROS and on operating systems used in the automotive industry.

\subsection{Robotics Software Middleware}

A comprehensive review of robotics software middleware can be found in~\cite{Chitic_2018}. The author argues that robots need to advertise their functionality as services in order to allow other members of the fleet to interact with them. The cited robotics middleware have been classified into four groups, based on their mechanisms used for communicating with distributed components:

\begin{itemize}
    \item Remote Procedure Call (RPC)~\cite{Birrell_1984} and Distributed Object Middleware (DOM) middleware~\cite{Capra_2001};
    \item Message-Oriented Middleware (MOM)~\cite{Curry_2004};
    \item Transaction-Oriented Middleware (TOM)~\cite{Issarny_2007};
    \item Service Oriented Architecture (SOA)~\cite{Papazoglou_2007}, Service Oriented Middleware (SOM)~\cite{Schmidt_2005_IBM} and Enterprise Service Bus (ESB)~\cite{Schmidt_2005_IBM}.
\end{itemize}

Based on the number of citations, the eight most common robot middleware are:

\begin{itemize}
    \item Robot Operating System (ROS)~\cite{Quigley_ROS_2009};
    \item Open Robot Control Software (OROCOS)~\cite{Borghesan_2020};
    \item Player/Stage~\cite{Kranz_2006};
    \item Nvidia Isaac SDK~\cite{Lowndes_2021};
    \item Mobile Robot Programming Toolkit (MRPT)~\cite{Claraco_2010};
    \item Webots~\cite{Norton_2019};
    \item Microsoft Robotics Developer Studio (MRDS)~\cite{Morgan_2019};
    \item Orca~\cite{Makarenko_2006}.
\end{itemize}

The \textit{Robot Operating System} (ROS)~\cite{Quigley_ROS_2009} is a collection of tools, libraries and conventions used to facilitate distributed communication over a wide variety of platforms. Since ROS is currently the most used middleware in robotics, we have dedicated subsection~\ref{sec:ros} for its analysis.

The \textit{Open Robot Control Software} (OROCOS)~\cite{Borghesan_2020} is a collection of portable C++ libraries for advanced machine and robot control. Its two main components are the Orocos Real-Time Toolkit (RTT), used for writing C++ modules, and the Orocos Component Library (OCL), used to start an application and interact with it at run-time.

\textit{Player/Stage}~\cite{Kranz_2006} was one of the first robotics middleware, having as goal the development of free software for robotics and sensor systems and also being a repository server for sensors and actuators. It also used to include the Gazebo simulator, which is now an independent software solution.

\textit{Nvidia Isaac SDK}~\cite{Lowndes_2021} is a toolkit that includes software building blocks for developing robots equipped with Nvidia hardware, having as focus perception and navigation using AI methods. It is not a middleware for distributed communication between multi-robot systems, but a collection of algorithms optimized for Nvidia processors, which uses ROS as a communication framework.

The \textit{Mobile Robot Programming Toolkit} (MRPT)~\cite{Claraco_2010} is a cross-platform library written in C++ for research in the area of Simultaneous Localization and Mapping (SLAM), computer vision and motion planning.

\textit{Webots}~\cite{Norton_2019} is not a direct hardware-interfaced robotics middleware, but a 3D robot simulation environment. It is based on the Open Dynamics Engine (ODE) for rigid body simulation and collision detection, allowing the precise simulation of the physical properties of objects, such as velocity, inertia and friction. An API is provided for interfacing with the actual robot controller, implemented outside Webots in C, C++, Python, ROS, Java, or Matlab.

\textit{Microsoft Robotics Developer Studio} (MRDS)~\cite{Morgan_2019} is a now discontinued Windows environment for robotics simulation and control. Communication is enabled using a .NET concurrent library implementation for managing asynchronous parallel tasks.

\textit{Orca}~\cite{Makarenko_2006} is an open-source framework for developing component-based robotic systems. It is designed to create building block that can be interconnected between each other, having code reusability as a central objective.

Although all the software middleware referenced above provide capabilities for distributed multi-robot communication, they lack in embedding the mechanisms required to build and deploy AI robotic components. AI modules are either encapsulated in packages, without any re-train and update functionalities, or the frameworks are patched with separated AI stacks. These stacks are typically created with design patterns that serve other types of applications, thus imposing a high overhead on their maintenance and stability. A notable example here is Nvidia Isaac, where ROS was patched with a separate SDK for enabling proper AI inference implementation.

These types of architectures allow solely for independent components development and testing, making it difficult to test the complete robotics chain. In contrast to this, the algorithms in CyberCortex.AI are developed and tested as an end2end application by evaluating the complete DataBlock during development.

A general purpose robotics operating system for unmanned vehicles (autonomous cars or drones) is disclosed in~\cite{Perrone_2017}. The disclosed system describes the software architecture for a single computer which runs the vehicle control system as a sequential processing pipeline without taking into account how different vehicles exchange data between them. Also, the invention does not describe how AI and learning systems can be designed, trained and deployed using the proposed approach. Similarly, a dynamic and collaborative design approach for automation and robotics software systems was proposed in~\cite{RoboticsSoftwareSystems}.

A hybrid robotic system application framework based on a multi-core processor framework is disclosed in~\cite{Wei_2014}. The invention presents how two operating systems (a real-time OS and a non-real-time OS) can be run on the same computing platform using multiple cores. The non-real-time OS uses a ROS node~\cite{Quigley_ROS_2009} for communication. As in~\cite{Perrone_2017}, the disclosure does not present a new communication system between multiple robots, as well as how training and deployment of AI components can be performed.

A method and device for managing shared memory in ROS is presented in~\cite{Feng_2016}. The invention describes how a shared memory accessed by a ROS node can be managed in a vehicle (car) controller. Such an additional system is required due to the faulty ROS design, which does not include memory management.

Methods and systems for multirobotic management are also disclosed in~\cite{Hickman_2012}, where multiple client robots are connected to a leader robot, while sharing a common database. Apart from the absence of functionalities for AI components, the approach lacks decentralization, being dependent on a central leader robot.

\subsection{Robot Operating System (ROS)}
\label{sec:ros}

The monopoly in robotics software is currently held by ROS (Robot Operating System)~\cite{Quigley_ROS_2009}. ROS itself is not an operating system in the traditional sense of process management and scheduling, but instead provides a communication layer between computing units. The goals of ROS, stated in~\cite{Quigley_ROS_2009}, are peer-to-peer communication, tool based, multi-lingual, thin and free open-source.

As reported by various research and industry groups, ROS is a bulky and difficult to use middleware, with a steep learning curve~\cite{PuluRobotics}. Typically, the ROS distributions require large computational resources and are mainly stable on Ubuntu Linux. Even though ROS has been released in 2009, Windows support has been included only in 2019. However, its main drawback is the lack of support in the design, development and evaluation of AI components, such as the deep neural networks used for perception and control. This is understandable, since ROS has been developed in the first decade of the 2000's, before the age of artificial intelligence and deep learning. In the following, we will tackle the drawbacks of ROS and how CyberCortex.AI overcomes these.

Although claimed that the principles of ROS allow for rapid prototyping of robotics systems, engineers have to deal with difficult integration problems and long module compilation times. This is particularly a problem in robotics applications which have a small to medium complexity, in the sense that ROS introduces a high level of overengineering within the overall application in general, as well as within the individual components. Having a light architecture structure in the form of a DataBlock, CyberCortex.AI can be compiled and deployed on small devices, such as Raspberry PI embedded computers. Due to the light DataBlock, the compilation times are low.

Peer-to-peer communication in ROS is implemented using the TCP/IP and UDP protocols\footnote{\url{https://tinyurl.com/4u62pjm2}}. This restricts the robotic systems and HPC cloud to run on the same LAN network. The development of additional solutions is required to enable communication between robots located in different LAN networks, such as routing the data over a VPN network. Additionally, the Data Distribution Service (DDS) protocol was added to ROS 2 as the main communication backbone, which provides a publish-subscribe transport which is very similar to ROS’s publish-subscribe transport, making ROS dependent on yet another software system. CyberCortex.AI uses a protocol agnostic DataChannel, which can be used using any communication backbone. In our research, we have implemented the DataChannel on top of the WebRTC protocol, where the information encoded in the DataBlock is directly transmitted over the internet. Since the DataChannel is just another filter in the DataBlock, it can be replaced by another DataChannel implemented on top of another protocol, such as Ethernet, or FlexRay.

Robotics applications in ROS are built by interconnecting modules implemented as publisher-subscriber ROS nodes, similar to how the Filters in CyberCortex.AI locally interact with each other. However, instead of manually connecting ROS nodes individually between them, we share the whole DataBlock's metadata via Signaling Servers, thus automating the discovery of Filters between robots.

Each ROS node is executed in its own process, node-to-node data transfer being achieved via inter-process communication. This requires the nodes to implement additional memory management mechanisms in order not to mistakenly overwrite the same memory space. This additional effort has been reported in the invention note~\cite{Feng_2016}, where the authors had to designed an extra mechanism for memory management. In contrast to this approach, each Filter running within the CyberCortex.AI's DataBlock is executed in is own thread, making our implementation compliant with the requirements of real-time operating systems, such as QNX or FreeRTOS, where each scheduled task is executed in a separate thread.

A key performance difference between our approach and ROS is also the local data transmission method. The most common method for data transfer in ROS is the TCPROS protocol, which is based on standard TCP/IP sockets for each ROS node. CyberCortex.AI starts multiple filters in the same process, thus allowing for in-memory data transfers based on our \textit{Temporal Addressable Memory} (TAM), thus achieving data transfer speeds as high as the memory supports.

The major advantage of ROS over other robotics software frameworks is the large international community supporting its development. This makes it possible to use already implemented ROS packages. However, using these packages comes at the cost of overengineering the applications. Although integrating the packages should be straightforward, the time required to adapt the hyperparameters and test the included algorithms is very often higher than just integrating the source code of an algorithm into a more lighter and scalable framework. This flexibility is provided in CyberCortex.AI through the scalability of the DataBlock, where each Filter can integrate algorithms provided as open-source software.

Although claimed in the design principles to be thin~\cite{Quigley_ROS_2009}, the ROS distribution has a large footprint ($>1$ GB), requiring a lot of hard-drive memory space for installation and high computational resources to run it. Additionally, even though it has been released in 2009 for Ubuntu Linux, ROS only added Windows support a decade later, in 2019. This time gap shows the difficulties of its scalability and portability. CyberCortex.AI has been from the start designed and tested on Linux, Windows and the Android operating system.

The main reason that prompted us to develop CyberCortex.AI is the lack of ROS mechanisms required to design, develop and test AI modules for robotics. Research and industry groups are either patching together ROS with external AI frameworks (e.g. ROS and Nvidia Isaac, or ROS and Microsoft.AI in Windows), or are developing their own AI robotics stacks. A prominent example is Tesla, which decided to develop its own Linux-based operating system capable of streaming large quantities of data to HPC clouds for training DNNs, while also implementing the inference engines required to process the data acquired from their cameras.

\subsection{Automotive Operating Systems}

Since the automotive domain is a subfield of robotics, it is worth visiting the operating systems and middleware implemented for such applications and their relation to AI technologies~\cite{Grigorescu_2020}.

Automotive grade OSs are divided into critical operating systems, used for controlling vital driving functionalities (e.g. cruise control, airbag status, etc.), and infotainment operating systems, used for running the car's HMI (Human Machine Interface). In both cases, the underling framework is a real-time OS, such as QNX or Linux.

Any software component developed for the automotive industry has to comply with the ISO 26262 standard for functional safety of road vehicles, describing four Automotive Safety Integrity Levels (ASIL)~\cite{Salay_2017}. ASIL D represent the highest degree of rigor required to reduce risk (e.g., testing techniques, types of documentation required, etc.). Initially defined in 2011, the ISO 26262 standard does not yet include functional safety methodologies for developing AI components for road vehicles. Safety and trustworthiness are requirements for AI-based operating systems for automotive applications, as well as for industrial software~\cite{IoT_AI_Safety}.

Different consortiums have been formed in order to assure scalability and software reusability between different vehicles. The most notable ones are \textit{AUTomotive Open System ARchitecture} (AUTOSAR)~\cite{Autosar_2020} and \textit{Automotive Grade Linux (AGL)}~\cite{Agl_2018}.

Although efforts are underway, they do not yet include any description of how AI and deep learning software should be developed, evaluated and maintained within the product lifecycle of a vehicle. The CyberCortex.AI architecture offers a complete description of how such a software can be designed and deployed, using training data directly acquired from fleets of cars.

The lack of standardization prompted companies to develop their own in-house solutions. Tesla for example, developed its Full Self-Driving (FSD) computer and software for implementing Advanced Driver Assistance Systems (ADAS) and autonomous driving capabilities, using a modified Ubuntu Linux distribution and without relying on any of the above cited robotic middleware.

\vspace{2mm}

In the light of current approaches and their limitations, we propose the CyberCortex.AI architecture, which aims to provide a holistic framework for developing robotics applications based on AI technologies.

\section{CyberCortex.AI Architecture}
\label{sec:architecture}

\begin{figure}
\centering
	\begin{center}
		\includegraphics[scale=1.0]{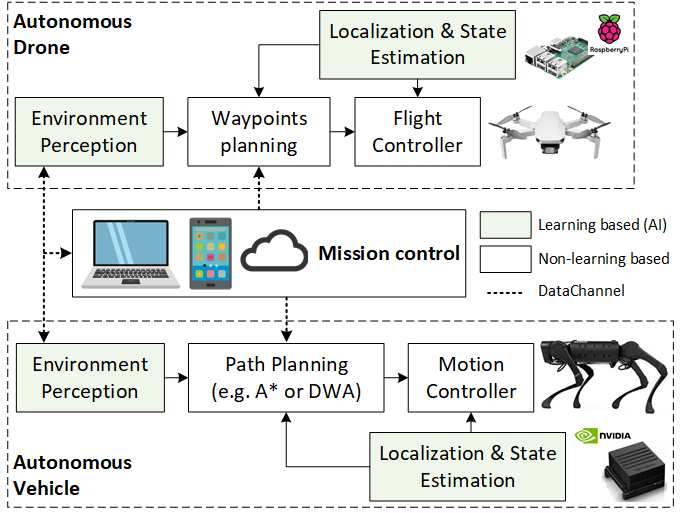}
        \vspace{-4mm}
        \caption{\textbf{Typical distributed robotics application in the field of autonomous surveillance.} The autonomous legged robot (UGV) and drone (UAV) can improve their path planners if they exchange perception data between them, as well as by sharing computational resources. A remote Mission Controller is used to pass high level commands, as well as to visualize and store sensory data. AI algorithms are mainly used in the perception, localization and state estimation components. The dotted lines indicate remote communication through the DataChannel.}
        \label{fig:block_diagram_robotics_pipeline}
	\end{center}
    \vspace{-8mm}
\end{figure}

The CyberCortex.AI architecture from Fig.~\ref{fig:block_diagram_simple} has been designed in order to enable the rapid development and maintenance of distributed robotics and complex automation applications based on AI. For the sake clarity, CyberCortex.AI will be presented through a distributed robotics application example in the field of surveillance for forest fires prevention, illustrated in Fig.~\ref{fig:block_diagram_robotics_pipeline}.

The hardware components used in this example are \textit{i}) a legged robot as an Unmanned Ground Vehicle (UGV) equipped with a powerful embedded computer (Nvidia AGX Xavier), LiDAR and cameras, \textit{ii}) an Unmanned Aerial Vehicle (UAV) equipped with a small computer for flight control (Raspberry PI) and \textit{iii}) a mission control system, where the user can upload geolocation data for the forest area to be surveilled. The mission control system can be located either in the cloud, directly on a laptop, or on a smartphone.

\begin{figure*}
\centering
	\begin{center}
		\includegraphics[scale=1.0]{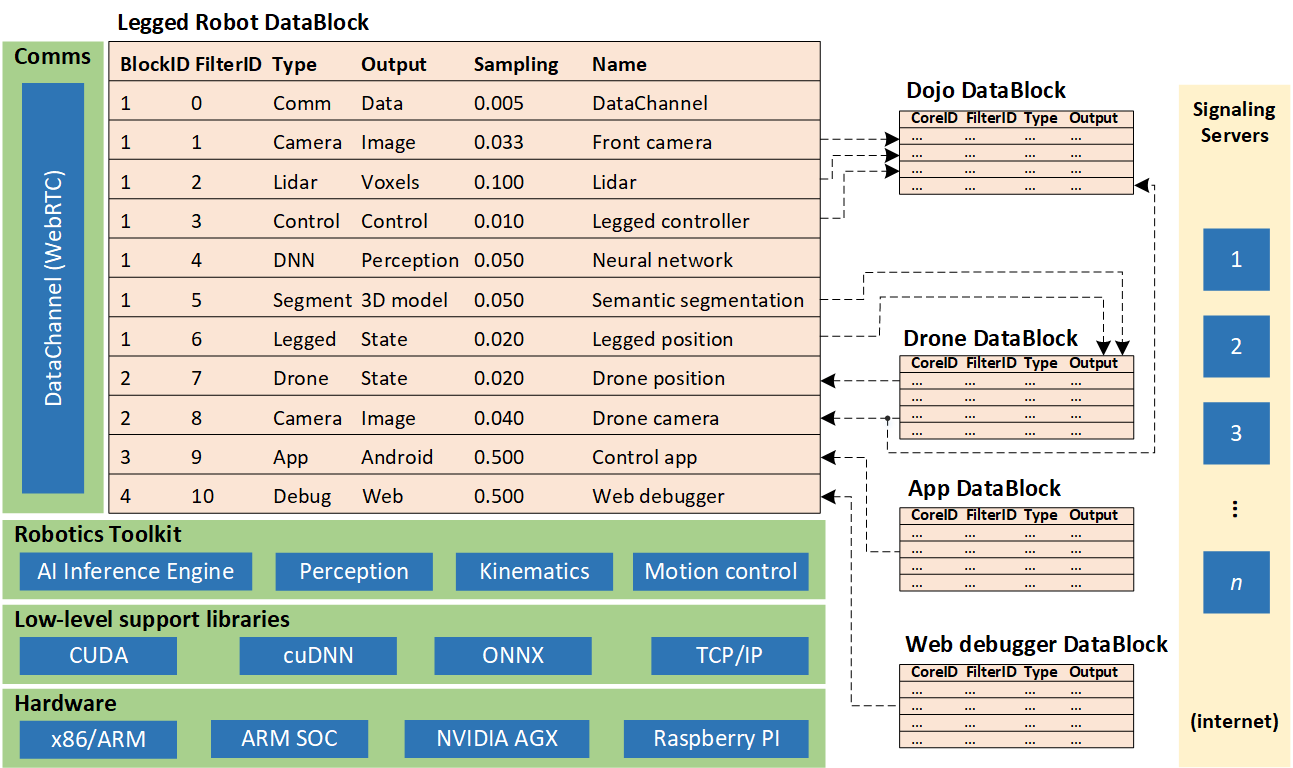}
        \vspace{-2mm}
        \caption{\textbf{Details of the CyberCortex.AI.inference system executed on an autonomous robot.} The communication between filters running on different DataBlocks is established using a redundant set of Signaling Servers, which act as a discovery system between the five DataBlocks (legged robot, drone, dojo, smartphone app and web debugger).}
        \label{fig:block_diagram_inference}
	\end{center}
    \vspace{-8mm}
\end{figure*}

Both autonomous robots are driven through perception-and-control pipelines, where the paths and waypoints are planned using data from the \textit{Environment Perception} and \textit{Localization \& State Estimation} modules. Motion and flight control is executed based on the calculated paths and waypoints, respectively.

The algorithms running in the \textit{Environment Perception} and \textit{Localization \& State Estimation} components from Fig.~\ref{fig:block_diagram_robotics_pipeline} are learning-based, meaning that saved sensory and control data is required for training their AI architectures prior to deployment. The AI architectures encountered in such robots are variants of Convolutional (CNN), Recurrent (RNN) and fully connected Neural Networks. Training is typically performed offline using powerful HPC cloud computers.

Apart from the local control system, each autonomous robot can benefit in its mission objectives by using, for example, environment perception information obtained by the other robot. The UGV can plan its path based on obstacles visible solely from the air, while the UAV can plan its waypoints using obstacles detectable only with the LiDAR installed on the UGV. Additionally, computational intensive operations, such as the forward pass through a large deep neural network, should be performed on the most powerful computing platform shared by the robots, that is, on the embedded computer installed on the legged robot. Both functionalities require a fast and stable communication between the two platforms.

CyberCortex.AI (see Fig.~\ref{fig:block_diagram_simple}) was designed in order to create AI-based, distributed and scalable robotics applications, such as the one described above. It provides the following features:

\begin{itemize}
    \item rapid implementation of perception-and-control pipelines through Filter components running in a so-called DataBlock; each filter output generates a datastream that can be used locally, as well as shared with other robots through the internet;
    \item a software architecture designed for processing sequences of temporal data, access through our \textit{Temporal Addressable Memory} (TAM);
    \item a stable real-time communication system which enables robots to automatically tap into remote datastreams and hardware resources; the automatic discovery of datastreams is enabled by registering a DataBlock into redundant Signaling Servers;
    \item a DataBlock having a small footprint ($<$ 5 MB), which is the CyberCortex.AI.inference system;
    \item a rapid design, training, deployment and maintenance platform for AI components, using datastreams channeled to a powerful HPC cloud system, that is, the CyberCortex.AI.dojo.
\end{itemize}

In the following, we describe the details of each main module in the CyberCortex.AI architecture.

\subsection{CyberCortex.AI.inference}

The CyberCortex.AI.inference system, illustrated in Fig.~\ref{fig:block_diagram_inference}, is a C/C++ implementation of the \textit{DataBlock}. We will detail its design principles based on the distributed robotics application from Fig.~\ref{fig:block_diagram_robotics_pipeline}.

Fig.~\ref{fig:block_diagram_inference} shows five DataBlocks running on five different devices. For the sake of clarity, only the DataBlock executed on the legged robot is detailed (left area of the figure). A DataBlock is a table-like structure specifying the structure of the processing pipeline running on a computing device that has an unique BlockID. The BlockID is equal to the device's MAC address. Each entry in the DataBlock is a so-called \textit{Filter} running at a given sampling time. A filter outputs a processed datastream, while consuming datastreams provided by other filters. The output datastream is obtained either by locally processing the input datastreams, or by mapping the output of a filter computed on a remote device. In the later case, the output datastream of a filter running on a remote computer is mirrored locally, thus masking the actual location of computation. Please refer to Section~\ref{sec:datachannel} for details on the communication and synchronization mechanisms.

In the example from Fig.~\ref{fig:block_diagram_inference}, the BlockID of the legged robot's DataBlock is 1. The first 6 filters on the legged robot are computed locally, while the rest are remote filters computed on the drone (BlockID 2), on a smartphone app (BlockID 3) and within a web browser (BlockID 4). Using this mechanism, remote computed datastreams are injected into the perception-and-control pipeline of the legged robot. Namely, the drone's camera and state are used as additional data for controlling the robot. The other way around, the same mechanism is used to transmit the perceptual information (e.g. traversable areas and/or detected objects) and the legged robot's position to the drone.

The example also illustrates how computing resources can be shared between robots. The datastreams from both the UGV's and the UAV's cameras are processed on the legged robot by the neural network filter 4, which receives as input the datastreams from filters 1 (UGV camera) and 2 (UAV camera). As detailed in Section~\ref{sec:ai_inference_engine}, the DNN outputs a datastream comprising of various prediction data, depending on the DNN architecture (e.g. object lists, semantic segmentation, road boundaries, etc.). The perception information is used locally on the legged robot for path planning, as well as on the drone for waypoints planning. In the diagram from Fig.~\ref{fig:block_diagram_inference}, the Semantic segmentation filter constructs a 3D octree model of the environment using the output of the DNN. The 3D model is shared between the UGV and the UAV for path and waypoints planning.

The design of a new filter is supported by a Robotics Toolkit that provides perception, kinematics and motion control functions, as well as the AI inference engine detailed in Section~\ref{sec:ai_inference_engine}). The perception functions include Simultaneous Localization and Mapping (SLAM), Visual-Inertial Odometry (VIO) pipelines and basic 3D reconstruction and keypoints tracking methods. The kinematics module includes different robot models (e.g legged robots, drones, wheeled vehicles and redundant manipulator models), while the motion control module provides simple PID control functions, as well as horizon and model predictive control methods that can be included in the DataBlock. The Robotis Toolkit is scalable and can be upgraded with additional methods.

Low-level libraries are included in order to support the development of the Robotics Toolkit functions. These include libraries for parallel processing on Graphical Processing Units (GPU), such as CUDA, libraries for propagating data through deep neural networks, such as cuDNN and ONNX~\cite{Le_2020}, as well as basic WebRTC or TCP/IP network communication functions.

CyberCortex.AI.inference has been successfully compiled on different operating systems (Linux, Windows and Android) with different types of processors, such as PC's equipped with x86 and ARM processors, ARM System-On-a-Chip (SOC) devices (smartphones), or Nvidia AGX Xavier and Raspberry PI embedded boards.

The disk size and resource utilization of CyberCortex.AI.inference depends on the number of filters included in the DataBlock. For example, a filter implementing a DNN forward pass requires both extra storage and computing resources, as opposed to a filter which just acquires data from a sensor. Without any complex filters included, the footprint of CyberCortex.AI.inference is small ($<$ 5MB), enabling its deployment on performant embedded boards and HPC clusters, as well as onto small embedded hardware (e.g. Raspberry PI).

\subsection{DataChannel and Temporal Addressable Memory (TAM)}
\label{sec:datachannel}

The remote connections used to map the filters between the DataBlocks in Fig.~\ref{fig:block_diagram_inference} are established using a so-called protocol agnostic DataChannel. In our work, we have implemented the DataChannel on top of the WebRTC protocol~\cite{Sredojev_2015}. WebRTC enables real-time communication capabilities by supporting video, voice and generic data to be sent between peers. In the same fashion, the DataChannel can be implemented using other protocols, such as Ethernet, of FlexRay.

\begin{figure}
\centering
	\begin{center}
		\includegraphics[scale=1.0]{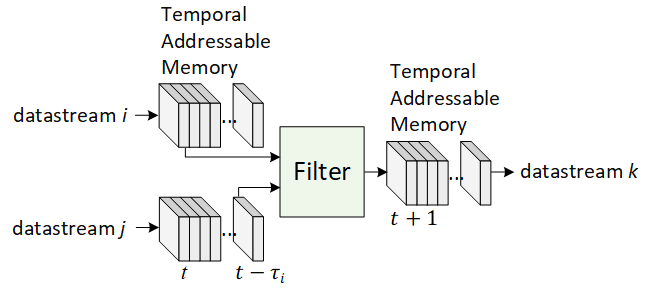}
        \vspace{-2mm}
        \caption{\textbf{Temporal Addressable Memory (TAM) used for data synchronization and propagation through the AI inference engine}. The output of each filter is stored in a finite FIFO cache memory indexed by timestamps. The synchronization of two Filters running at different sampling rates is performed by addressing the TAM with the desired timestamp.}
        \label{fig:temporal_memory}
	\end{center}
    \vspace{-8mm}
\end{figure}

A redundant set of Signaling Servers are used to establish peer-to-peer communication over the internet, that is, the communication required by datastreams connecting mapped filters running on different DataBlocks. A signaling server's sole purpose is to setup the peer-to-peer communication.

When a DataBlock is started, it automatically sends its clock time and metadata (BlockID, filter IDs, input/output types, sampling time and description of each filter) to $n$ signaling servers. This allows a robot to discover the location and resources of other robotic systems.

In theory, a single signaling server can be used to establish peer-to-peer communication for all robots. However, if that single signaling server stops functioning, then the robots will not be able to start the peer-to-peer datastreams. By using $n$ redundant servers, distributed worldwide, we minimize the risk of not being able to grant peer-to-peer connections.

The \textit{Temporal Addressable Memory} (TAM) from Fig.~\ref{fig:temporal_memory} has been implemented in order to synchronize different datastreams, as well as to manage the data propagation process through our AI inference engine. It acts as a First-In-First-Out (FIFO) cache of size $\tau_i$, which stores a finite number of filter output samples together with their timestamps. The TAM is indexed and addressed by timestamps. These timestamps can be used to synchronize data coming from two filters running at different sampling rates. The filter from Fig.~\ref{fig:temporal_memory} uses datastreams $i$ and $j$ as input, while delivering datastream $k$ as output. The synchronized samples are read from the TAM using the available timestamps.

Apart from the C/C++ implementation, a Javascript variant of the DataBlock is also available for debugging purposes on our website\footnote{\url{https://cybercortex.ai/login.php}}. Using a browser, one can tap into a Signaling Server to check and visualize the outputs of the registered DataBlocks on that specific server. The web visualizer and debugger is a javascript implementation of the DataBlock executed within a browser, which enables one to connect to a given Signalig Server and tap into the registered DataBlocks. It is used to visualize the filters' outputs, update the DataBlock and filters to new versions, as well as to pass high-level mission commands.

\begin{figure}
\centering
	\begin{center}
		\includegraphics[scale=1.0]{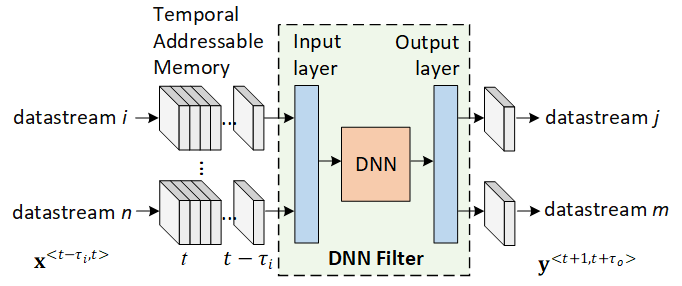}
        \vspace{-5mm}
        \caption{\textbf{The AI inference engine} organizes its input and output branches as temporal sequences of data from the Temporal Addressable Memory (TAM), indexed through timestamps. The inputs are extracted from the TAM, while the outputs are stored in the TAM.}
        \label{fig:dnn_inference_engine}
	\end{center}
    \vspace{-8mm}
\end{figure}

\subsection{AI Inference Engine}
\label{sec:ai_inference_engine}

One of the main components in the Robotics Toolkit is the AI Inference Engine illustrated in Fig.~\ref{fig:dnn_inference_engine}. It is used for passing input datastreams through the layers of a Deep Neural Network (DNN). The architecture of the inference engine has been designed in conjunction with the CyberCortex.AI.dojo system, which is used for designing and training DNNs. Namely, DNNs constructed and trained on our dojo are automatically sent to the DNN filters in the registered DataBlocks.

The philosophy of the AI inference engine is based on our previous work from~\cite{Grigorescu_2021}. Temporal sequences of multiple input datastreams $\vec{x}^{<t-\tau_i, t>}$ are used to produce temporal sequences of output datastreams $\vec{y}^{<t+1, t+\tau_o>}$:

\begin{equation}
    \vec{x}^{<t-\tau_i, t>}=[\vec{x}^{<t-\tau_i>}, ..., \vec{x}^{<t>}],
\end{equation}

\begin{equation}
    \vec{y}^{<t+1, t+\tau_o>} = [\vec{y}^{<t+1>}, \vec{y}^{<t+2>}, ..., \vec{y}^{<t+\tau_o>}],
\end{equation}

\noindent where $t$ is the discrete timestamp, $\tau_i$ is the temporal length of the input samples and $\tau_o$ is the temporal prediction horizon. The elements of $\vec{x}^{<t-\tau_i, t>}$ represent cached data from the TAM.

Each datastream is passed to the input layer of the DNN as a separate input branch of variable batch size. Similarly, the output layer of the DNN provides datastreams through its output branches. The Open Neural Network eXchange (ONNX)~\cite{Le_2020} format was chosen for storing the structure and weights of the developed neural networks.

\subsection{DNNs Design, Training and Maintenance Pipeline}

\begin{figure}
\centering
	\begin{center}
		\includegraphics[scale=0.85]{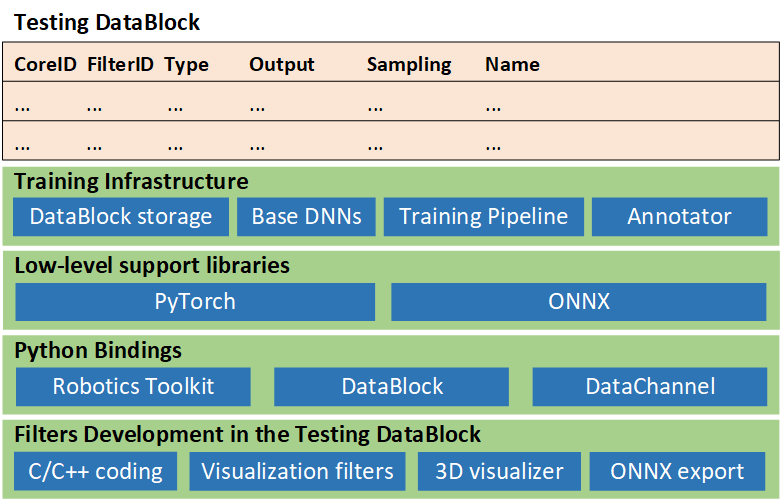}
        \vspace{-2mm}
        \caption{\textbf{CyberCortex.AI.dojo} used for storing information from DataBlocks, as well as for designing, training, deploying and maintaining DataBlocks and their AI components. It is a cloud-based implementation that can be executed on powerful HPC clusters.}
        \label{fig:block_diagram_dojo}
	\end{center}
    \vspace{-8mm}
\end{figure}

The design, implementation, training and maintenance of the filters within a DataBlock is enabled by the CyberCortex.AI.dojo system from Fig.~\ref{fig:block_diagram_dojo}. The dojo is mainly written in Python and executed on dedicated computers, such as powerful HPC clusters. It is composed of a development and testing DataBlock and a set of libraries which can be used for the rapid prototyping of the filters.

The low-level AI libraries of choice for designing and training DNNs are PyTorch~\cite{Paszke_2019} and ONNX~\cite{Le_2020}. PyTorch has been chosen due to its flexibility and support for vision based neural networks, while ONNX has been chosen for its open-source format, stability and portability.

A common base DNN structure has been implemented on top of PyTorch using the AI inference engine described in Section~\ref{sec:ai_inference_engine}. Each DNN exported from the dojo is inherited from this base structure. A DNN architecture is given in a configuration file which describes its input/output branches and tensor shapes. The same configuration file is loaded by the DNN filters in the CyberCortex.AI.inference system.

The training pipeline works by reading stored datastreams acquired from various DataBlocks running on robotic systems. In the example from Fig.~\ref{fig:block_diagram_inference}, the Dojo DataBlock stores the camera, LiDAR and control signals from the legged robot DataBlock, as well as the drone's camera. These are later used for training the perception DNN.

\begin{figure*}
\centering
	\begin{center}
		\includegraphics[scale=0.44]{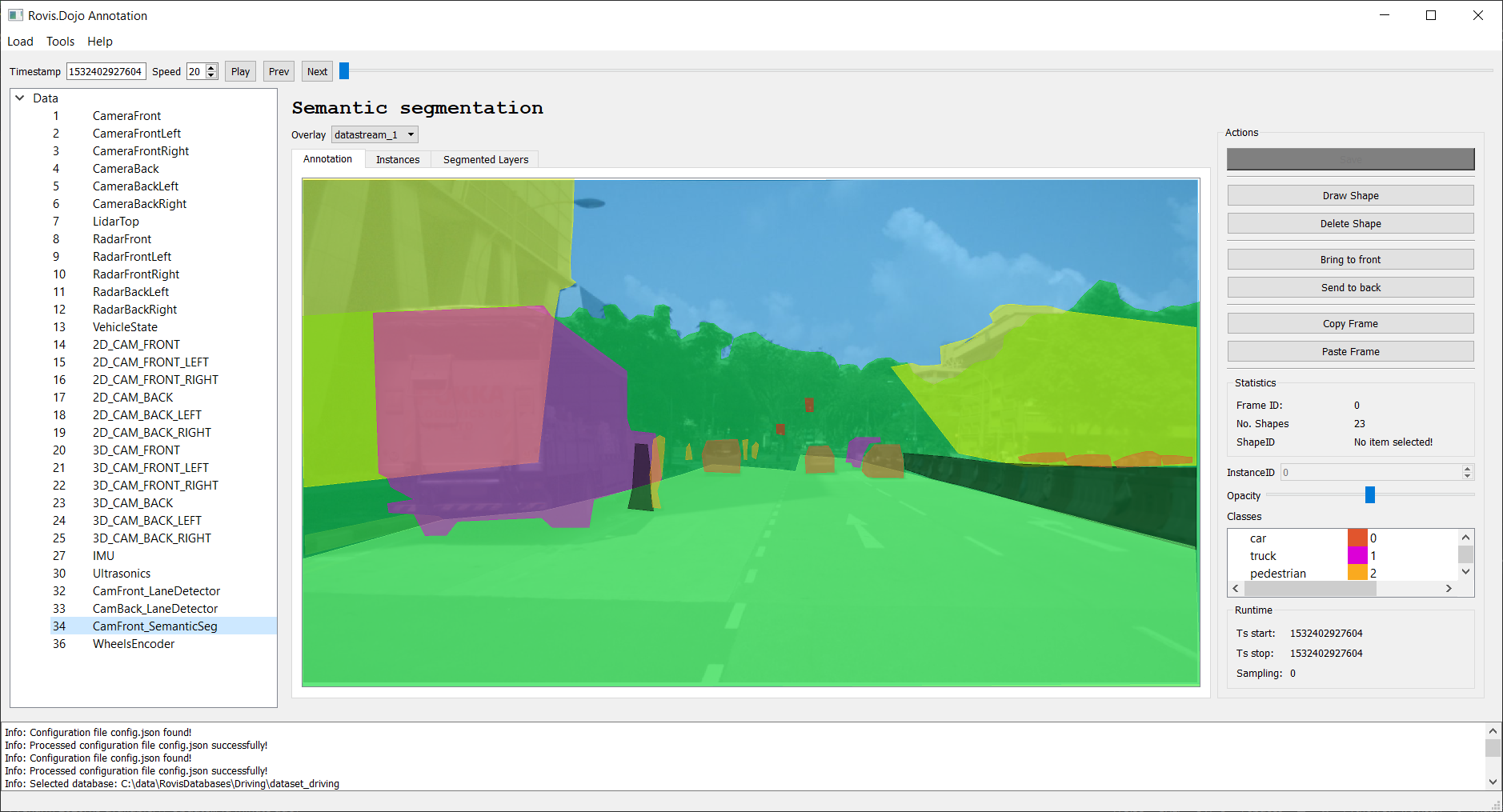}
        \vspace{-6mm}
        \caption{\textbf{Snapshot of the CyberCortex.AI Annotator tool}. The Annotator is used for the visualization and annotation of datastreams. The example shows a manually annotated segmentation datastream superimposed on the raw image datastream.}
        \label{fig:annotator}
	\end{center}
\end{figure*}

A neural network is trained by evaluating pairs of data samples $\vec{x}^{<t-\tau_i, t>}$ and labels $\vec{y}^{<t+1, t+\tau_o>}$. CyberCortex.AI treats both the samples and the labels as different datastreams. $\vec{x}^{<t-\tau_i, t>}$ are input datastreams, while the labels $\vec{y}^{<t+1, t+\tau_o>}$ are output datastreams. An Annotator is provided for visualizing the recorded datastreams. Fig.~\ref{fig:annotator} shows a snapshot of the Annotator during the analysis of a saved DataBlock. The recorded datastreams are shown on the left side of the tool, while the center area shows annotated regions of interest corresponding to the objects present in the scene.

The labels can be obtained either by manually annotating the corresponding datastreams using the Annotator tool, or by generating them using self-supervised learning methods.

The inner modules of a deep neural network are typically composed of trainable layers having different hyperparameters (convolutional and recurrent layers, ResNet and RegNet submodules, etc.). However, extensive analysis and work has to be done on the pre-processing and post-processing layers. For example, one-shot object detectors such as RetinaNet~\cite{Lin_2017} or Yolo~\cite{Redmon_2018} require object anchoring mechanisms and non-maximum suppression functions in order to recognize objects. These additional processing steps can be implemented using C/C++ functions from the Robotics Toolkit, which are also available in Python through a binding package. Once the DNN has been deployed, the functions can be called through their native C/C++ interface.

CyberCortex.AI.dojo already contains a number of filters for data acquisition, AI inference, perception, planning and control. These are written in C/C++ and can be used to construct a DataBlock. When a new filter is developed, it is tested by running the complete DataBlock of filters, as opposed to testing it as a separate component. This enables the capturing of the intrinsic dependencies between the whole components that make up a processing pipeline, instead of just optimizing a single component.

A special type of filters, mainly used during development, are the visualization filters. They consume input datastreams, while outputting images which superimpose the received inputs. For example, the sensing visualizer filter superimposes perception datastreams such as images, lidar, detected objects, keypoints and segmented surfaces. The dojo also includes a 3D visualization filter used in the analysis of perception systems and kinematic chains. A snapshot of the 3D visualizer can be seen in Fig.~\ref{fig:use_case_surveillance_deployment}(b). The visualization filters are demonstrated in the interactive demos described in Appendix~\ref{appendix:interactive_demo}.

\section{Performance Evaluation}
\label{sec:evaluation}

We have evaluated the performance of CyberCortex.AI by setting up two exemplary robotics use-cases: \textit{i}) the forest fires prevention scenario described in Section~\ref{sec:architecture} and \textit{ii}) an autonomous driving pipeline which uses CyberCortex.AI for collaborative perception. In the first use-case, we perform a quantitative evaluation of the framework by measuring computation and data streaming metrics, while in the second use-case, we analyze the impact of these metrics on the overall performance of the end2end control pipeline.

Both use-cases require trained and maintained AI components for perception, as well as real-time data communication and computational resources sharing. Since the current de-facto standard in robotics software is ROS~\cite{Quigley_ROS_2009}, we have evaluated CyberCortex.AI against equivalent ROS implementations of the two considered use-cases. Namely, we have implemented an equivalent ROS node for each filter in the CyberCortex.AI DataBlock.

\subsection{Forest Fires Prevention}

\begin{figure}
\centering
	\begin{center}
		\includegraphics[scale=0.78]{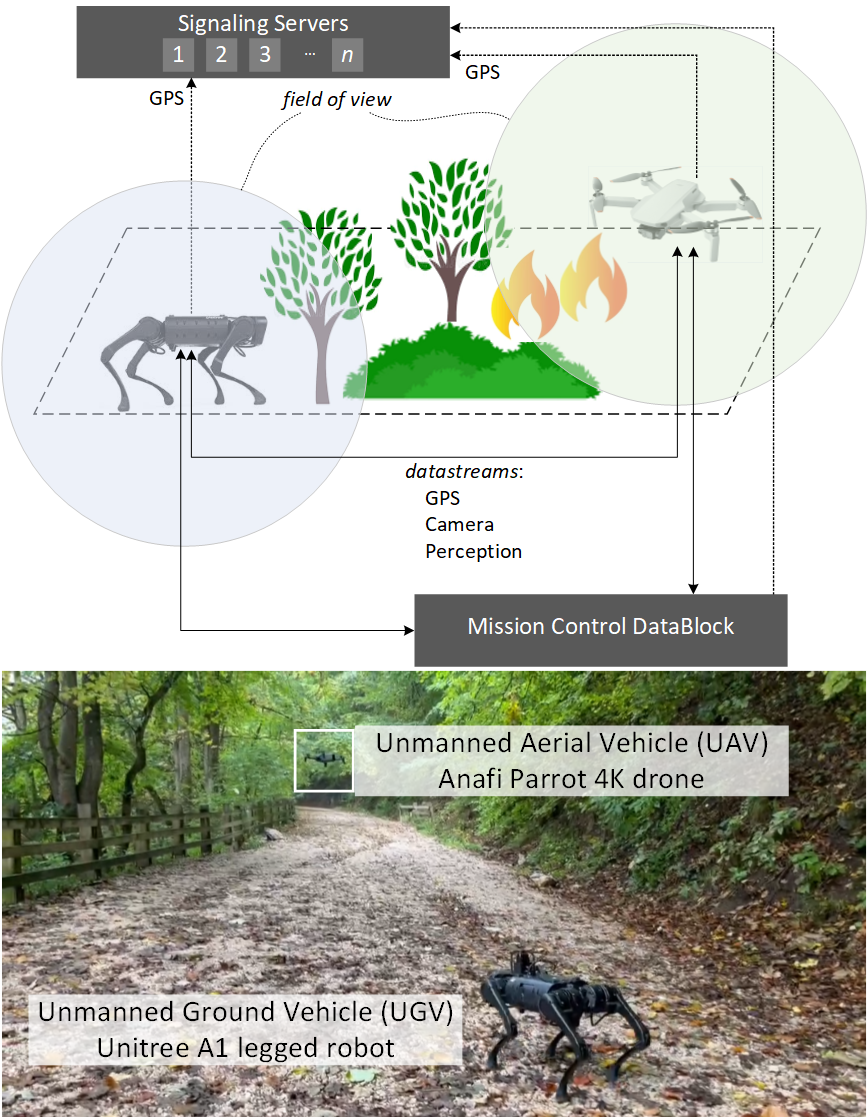}
        \vspace{-6mm}
        \caption{\textbf{Application of CyberCortex.AI to forest fires prevention.} The legged robot's (UGV) environment perception system is enhanced by the drone's (UAV) large field of view. The information collected by the drone is used by the legged robot for planning its intervention. Additionally, the data collected by both robots is sent to a Mission Control Server which decides on high-level actions.}
        \label{fig:use_case_surveillance}
	\end{center}
    \vspace{-3mm}
\end{figure}

The first experimental use-case, illustrated in Fig.~\ref{fig:use_case_surveillance}, is the forest fires prevention system from Section~\ref{sec:architecture}. The application is composed of an Unitree A1 legged robot, as the Unmanned Ground Vehicle (UGV), and an Anafi Parrot 4K drone as the Unmanned Aerial Vehicle (UAV). The two robotic systems receive high-level commands from Mission Control. These commands represent the perimeter of the area to be surveilled, which is typically located on unstructured rough terrain. The circular areas in Fig.~\ref{fig:use_case_surveillance} represent the field of view of each robot. Due to its larger size, the legged robot can be equipped with firefighting systems. However, the drone is solely used for scouting and information retrieval.

\begin{figure}
\centering
	\begin{center}
		\includegraphics[scale=0.72]{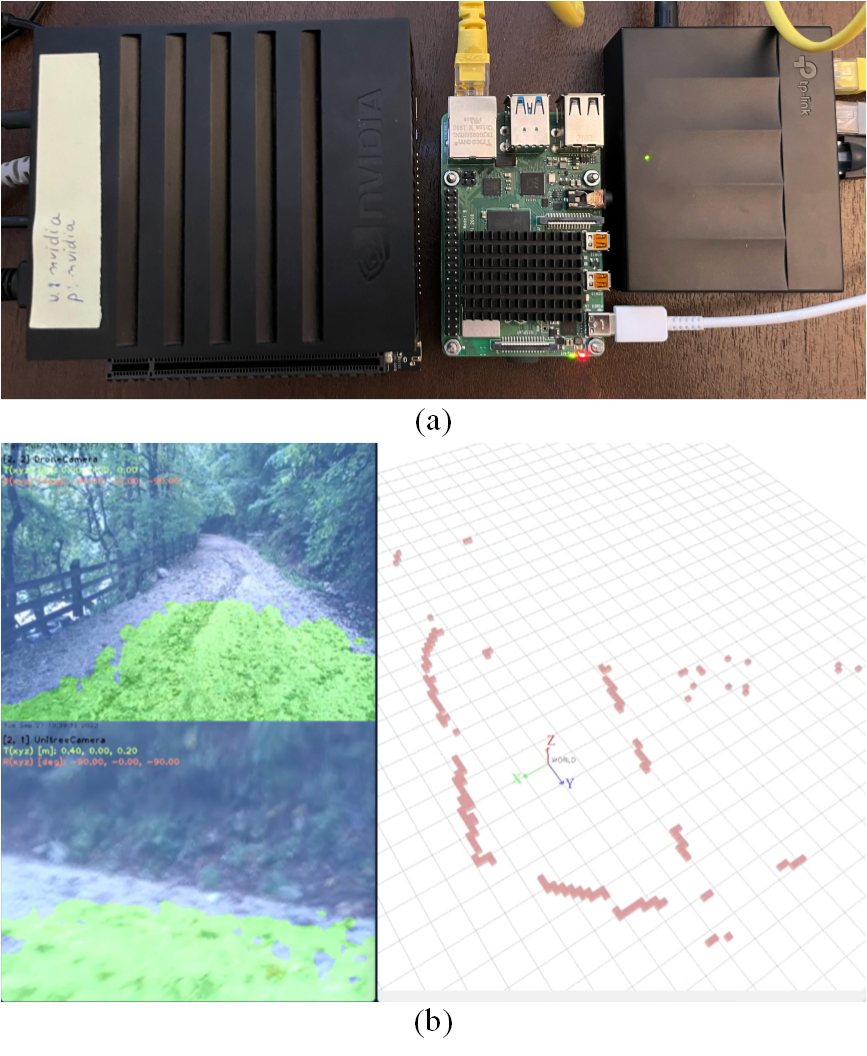}
        \vspace{-2mm}
        \caption{\textbf{CyberCortex.AI runtime environment.} (a) Hardware system: the Nvidia AGX Xavier board and Raspberry PI computer are used on the Unitree A1 legged robot and Anafi Parrot 4K drone, respectively. (b) Snapshot from video cameras superimposed on the semantic segmentation result, alongside the 3D environment model illustrating acquired 2D LiDAR data from the Unitree A1 robot.}
        \label{fig:use_case_surveillance_deployment}
	\end{center}
    \vspace{-6mm}
\end{figure}

The perception data represents the segmented traversable area calculated using a ResNet-50 encoder-decoder neural network~\cite{deeplabv32018}. Fig.~\ref{fig:use_case_surveillance_deployment} shows the hardware components used in the experiments, alongside superimposed semantic segmentation data and 2D LiDAR information. The Unitree A1 robot is equipped with the same embedded Nvidia AGX Xavier board as the one depicted in Fig.~\ref{fig:use_case_surveillance_deployment}(a). However, the Anafi Parrot 4K drone does not allow access to its underlining microcontroller for installing any additional software, such as CyberCortex.AI. In order to performed the experiments, we have instantiated CyberCortex.AI and ROS on the Raspberry PI computer from Fig.~\ref{fig:use_case_surveillance_deployment}(a), which is wirelessly connected to the Anafi drone for readings the camera and state information.

\begin{figure*}
\centering
	\begin{center}
		\includegraphics[scale=1.0]{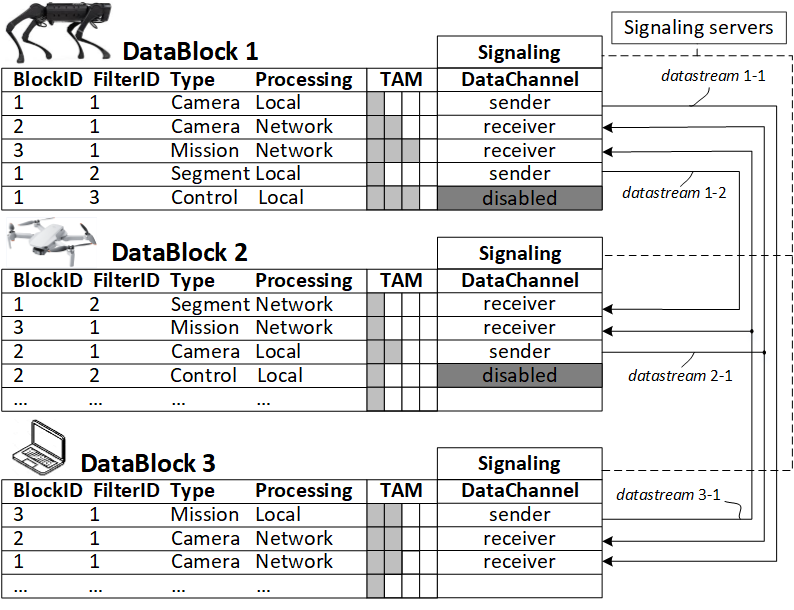}
        \vspace{-4mm}
        \caption{\textbf{Dataflow between the robotic systems and the mission control computer, respectively.} The dotted lines connecting the three systems with the Signaling Servers are used only during the initialization phase for establishing the datastreams connections depicted with solid lines. The datastreams are used to send the data stored in the Temporal Addressable Memory (TAM) to the respective peers.}
        \label{fig:flowchart_DataChannel}
	\end{center}
    \vspace{-6mm}
\end{figure*}

Fig.~\ref{fig:flowchart_DataChannel} illustrates the initialization and flow of data during operation. The connections between the three systems are established via the Signaling Servers. Once the robots have discovered themselves using the servers, they can directly exchange position, sensory data, perception information and computational resources.

Considering the performance of the end2end pipeline, we evaluate the computational load, communication overhead, memory footprint and booting time in the form of latency. The communication latency is the time required by a component (CyberCortex.AI filter, or ROS node) to interact with the communication layer of the operating system and transmit a message to another component. The latency is defined as:

\begin{equation}
    L_{t, t+1} = T_{t+1, R} - T_{t, S},
    \label{eq:latency}
\end{equation}

\noindent where $L_{t, t+1}$ is the latency, while $T_{t+1, R}$ and $T_{t, S}$ are the receiver and sender timestamps.

\begin{figure*}
\centering
	\begin{center}
		\includegraphics[scale=0.9]{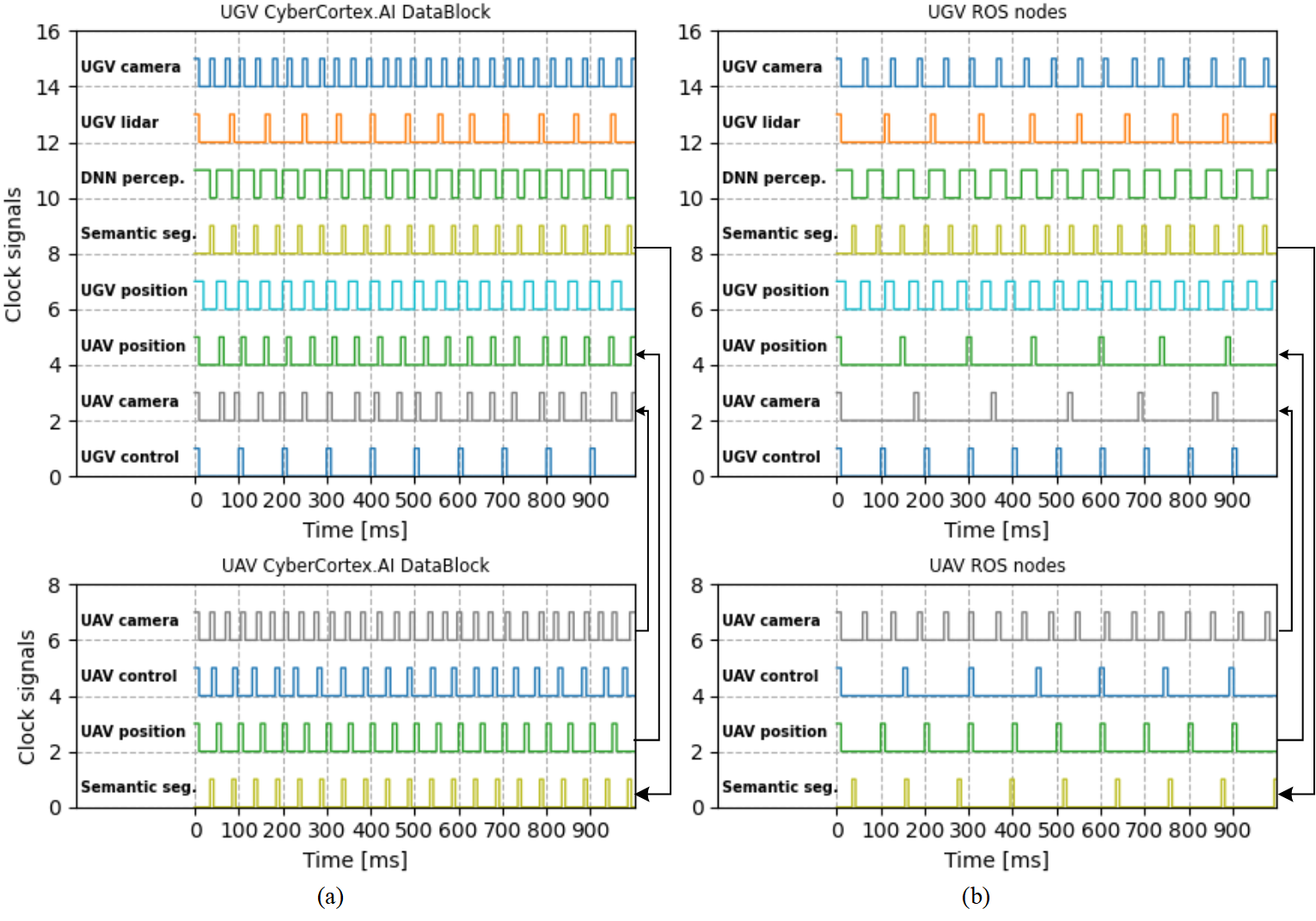}
        \vspace{-8mm}
        \caption{\textbf{Clock signals illustrating the sampling rate of the CyberCortex.AI DataBlock filters (a) and ROS nodes (b).} Each clock signal indicates when a filter, or ROS node, is computing data. Within CyberCortex.AI, the computed output data is stored in the Temporal Addressable Memory (TAM) during the clock signal's transition from the high to the low state. The arrows indicate the transmission of a datastream from one robot to another. Namely, the UAV is sending camera and positioning data to the UGV, while the UGV is sending semantic segmentation data to the UAV.}
        \label{fig:clock_signals_visio}
	\end{center}
    \vspace{-8mm}
\end{figure*}

The sampling rate of the CyberCortex.AI and ROS control pipelines can be visualized in Fig.~\ref{fig:clock_signals_visio}, for both the UGV and the UAV robotic systems. When active, a clock signal represents the computing cycle duration of a specific filter, where its high state indicates that the filter is processing data. The output data for each filter is stored in the Temporal Addressable Memory (TAM) during the clock signal's transition from the high to the low state. The arrows on the right side of the graphs indicate remote input/output datastream connections between the UGV and the UAV. The drone is sending camera and position data to the UGV, while the UGV processes the received data and sends back semantic segmentation information. In this particular example, the UAV camera and UAV control filters are considered remote filters on the UGV, since their actual computation takes place on the UAV, while the Semantic Segmentation filter, executed on the UGV, is considered a remote filter on the UAV. The remote filters are activated once their corresponding data is received through the DataChannel.

Fig.~\ref{fig:clock_signals_visio}(a) shows the filters running within the CyberCortex.AI DataBlock, while Fig.~\ref{fig:clock_signals_visio}(b) illustrates their ROS counterparts. To establish a proper comparison, both pipelines are running the same algorithms, encapsulated in the CyberCortex.AI filters and ROS nodes, respectively. The clock signals show the latency and computational overhead introduced solely by the two robotic operating systems. As visible in the diagrams, the sampling rate of CyberCortex.AI is a higher for the majority of the components, thus enabling the development of robotics applications composed of subsystems running at a higher frequency rate.

\begin{figure}
\centering
	\begin{center}
		\includegraphics[scale=0.76]{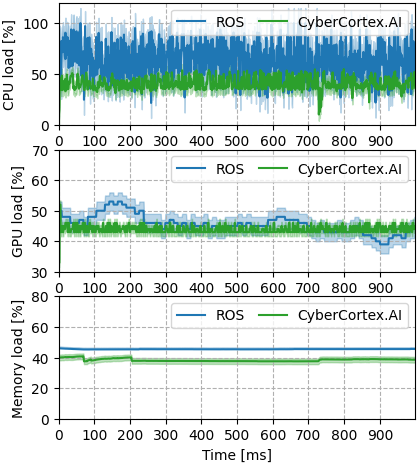}
        \vspace{-4mm}
        \caption{\textbf{Computational load on the UGV}, over a period of $1s$. The solid lines show runtime resources allocation on the legged robot's Nvidia AGX Xavier embedded board, while the shaded regions indicate the standard deviation.}
        \label{fig:computational_load_ugv}
	\end{center}
    \vspace{-8mm}
\end{figure}

\begin{figure}
\centering
	\begin{center}
		\includegraphics[scale=0.76]{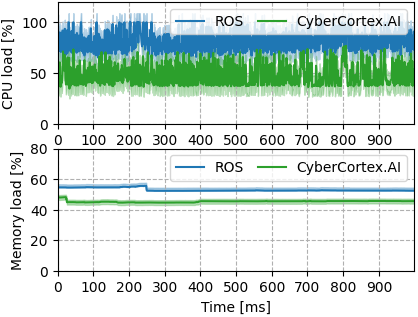}
        \vspace{-4mm}
        \caption{\textbf{Computational load on the UAV}, over a period of $1s$. The solid lines show runtime resources allocation on the drone's Raspberry PI embedded computer, while the shaded regions indicate the standard deviation.}
        \label{fig:computational_load_uav}
	\end{center}
    \vspace{-4mm}
\end{figure}

\begin{figure}
\centering
	\begin{center}
		\includegraphics[scale=0.74]{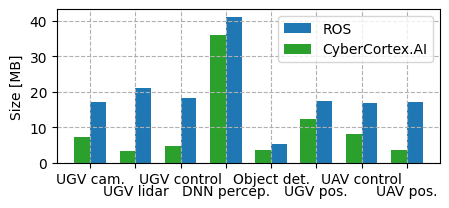}
        \vspace{-4mm}
        \caption{\textbf{Storage requirements for CyberCortex.AI's DataBlock filters and ROS nodes, respectively.} The graph shows the size of the compiled binary files.}
        \label{fig:storage}
	\end{center}
    \vspace{-8mm}
\end{figure}

A comparison of the computational load, memory requirements and storage demand, for both the UGV and UAV, can be seen in Fig.~\ref{fig:computational_load_ugv}, Fig.~\ref{fig:computational_load_ugv} and Fig.~\ref{fig:storage}, respectively. All the graphics show the CPU/GPU load and memory consumption of the considered processes, that is, of CyberCortex.AI and their corresponding ROS nodes.

The computational load data has been recorded in real-time, using the POSIX Kernel API for retrieving the CPU and memory consumption and the Nvidia Management Library (NVML) API for obtaining the state of the Nvidia AGX Xavier GPU. Because the ROS nodes are independent executables running in different processes, we have calculated their resources requirements by summing up their CPU/GPU load and memory allocation. Since the onboard computer of the Anafi Parrot 4K drone is not accessible, we have streamed the drone's camera and position data to a Raspberry PI computer running CyberCortex.AI and ROS. Considering that the Raspberry PI computer does not have a GPU, the graphics in Fig.~\ref{fig:computational_load_ugv} only show the CPU's computational load and memory requirements.

Both the UGV and the UAV require more CPU time when running the ROS nodes, as opposed to the CyberCortex.AI control pipeline. Additionally, the standard deviation is higher in the ROS nodes case, thus indicating an additional stress created during CPU time allocation.

\begin{table*}
	\centering
	\begin{tabular}{lcccccc}
		\hline
		\textbf{ROS node /} & \multirow{ 2}{*}{\textbf{OS}} & \textbf{Latency}& \textbf{CPU/GPU} & \textbf{Memory} & \textbf{Storage} & \textbf{Booting} \\
        \textbf{CyberCortex.AI Filter (CCR)} & & \textbf{[ms]}& \textbf{load [\%]} & \textbf{footprint [MB]} & \textbf{footprint [MB]} & \textbf{time [ms]} \\
        \hline
        \multirow{ 2}{*}{UGV Camera [UGV]} & ROS & - & 2.66 & 47.4 & 17.05 & 94 \\
        \cline{2-7}
        & CCR & - & \textbf{1.73} & \textbf{33.28} & \textbf{7.37} & \textbf{59} \\
        \hline
        \multirow{ 2}{*}{UGV Lidar [UGV]} & ROS & - & 2.57 & 45.68 & 21.08 & 127 \\
        \cline{2-7}
        & CCR & - & \textbf{1.08} & \textbf{38.72} & \textbf{3.28} & \textbf{68} \\
        \hline
        \multirow{ 2}{*}{DNN Perception [UGV]} & ROS & - & 49.64 & 107.12 & 41.27 & 527 \\
        \cline{2-7}
        & CCR & - & \textbf{43.91} & \textbf{101.76} & \textbf{35.9}1 & \textbf{482} \\
        \hline
        \multirow{ 2}{*}{Semantic segmentation [UGV]} & ROS & 29 & 2.70 & 26.76 & 5.39 & \textbf{91} \\
        \cline{2-7}
        & CCR & \textbf{16} & \textbf{0.91} & \textbf{23.68} & \textbf{3.66} & 163 \\
        \hline
        \multirow{ 2}{*}{UGV Positioning [UGV]} & ROS & - & 17.73 & 187.2 & 17.34 & 106 \\
        \cline{2-7}
        & CCR & - & \textbf{13.39} & \textbf{180.16} & \textbf{12.30} & \textbf{39} \\
        \hline
        \multirow{ 2}{*}{UAV Positioning [UAV]} & ROS & 21 & 2.95 & 22.08 & 17.19 & \textbf{85} \\
        \cline{2-7}
        & CCR & \textbf{14} & \textbf{1.05} & \textbf{19.52} & \textbf{3.7} & 138 \\
        \hline
        \multirow{ 2}{*}{UAV Camera [UAV]} & ROS & 31 & 2.63 & 26.32 & 16.73 & \textbf{101} \\
        \cline{2-7}
        & CCR & \textbf{23} & \textbf{0.98} & \textbf{21.76} & \textbf{8.02} & 193 \\
        \hline
        \multirow{ 2}{*}{UGV Controller [UGV]} & ROS & - & 19.92 & 237.2 & 18.14 & 99 \\
        \cline{2-7}
        & CCR & - & \textbf{13.72} & \textbf{232.64} & \textbf{4.75} & \textbf{37} \\
        \hline
        \multirow{ 2}{*}{\textbf{End2End pipeline} [UGV]} & ROS & - & 63.53 & 699.76 & 154.19 & - \\
        \cline{2-7}
        & CCR & - & \textbf{40.87} & \textbf{651.52} & \textbf{78.99} & - \\
		\hline
	\end{tabular}
	\caption{\textbf{Quantitative evaluation of latency and computational load}. Comparison of latency, CPU/GPU load, memory/storage footprints and booting time of the CyberCortex.AI DataBlock and ROS nodes, respectively.}
	\label{tab:quantitative_comparison}
    \vspace{-9mm}
\end{table*}

The GPU is allocated every time new input data is given to the AI inference engine for forward propagation. As shown in the middle diagram from Fig.~\ref{fig:computational_load_ugv}, the GPU allocation on the Nvidia board is relatively similar for both frameworks. This is mainly because neither CyberCortex.AI, nor ROS, require GPU based parallel computations in their respective core operating system. However, it is interesting to notice that the GPU gets allocated more frequently in the case of CyberCortex.AI. This phenomenon, visible in the GPU load from Fig.~\ref{fig:computational_load_ugv}, occurs due to the higher sampling rate of CyberCortex.AI, where the AI inference engine receives input data more frequently than in the case of ROS.

The lightweight structure of CyberCortex.AI is also noticeable through its memory consumption and disk storage requirements, the later illustrated in Fig.~\ref{fig:storage}. The sizes of the ROS components were calculated for each ROS node, while each compiled DataBlock filter was considered for the case of CyberCortex.AI. In particular, the size of the DNN perception component is large in both cases, since both CyberCortex.AI and ROS make use of the same deep neural network for processing visual data.

Table~\ref{tab:quantitative_comparison} summarizes the average latency, average CPU and GPU load, average memory and storage footprint, as well as the average booting time of the CyberCortex.AI filters and ROS nodes, respectively. The total booting time has not been summed up for the End2End pipeline, since each component is started in parallel, in its own thread, or process, respectively. The performance of the proposed framework is visible over all the considered metrics, making CyberCortex.AI more suited for real-time robotics applications over a large number of embedded computers.

The latency, computed using Eq.~\ref{eq:latency}, was calculated only for the remote filters. It shows the time required by a data package to be transferred from the sender to its destination filter, using the DataChannel in the case of CyberCortex.AI and the Data Distributed Service (DDS) in the case of ROS.

Apart from the higher transfer speed achieved using CyberCortex.AI, it is important to notice that ROS requires the communicating nodes to run in the same Local Area Network (LAN). This is opposed to CyberCortex.AI, which can use as the communication medium both the internet or the LAN. By restricting communication only to LAN networks, the robotics engineers are forced to create complicated port forwarding mechanisms for dealing with real-time data transfer. Since the DataChannel is not confined to a LAN network, it requires the establishment of a connection to the Signaling Servers, thus making the booting time for the remote filters higher than in the case of ROS. However, this latency is strictly linked to the booting time and does not affect the real-time communication between the filters.

\subsection{Autonomous Driving}

Additional to the quantitative study presented in the previous section, we have also performed a qualitative analysis of the proposed operating system using a collaborative autonomous driving setup. This analysis is not focused on the performance of the control algorithm itself, but on the implications of the operating system on the overall control pipeline. As in the previous section, we use the same algorithms within the CyberCortex.AI filters, as in the corresponding ROS nodes.

\begin{figure}
\centering
	\begin{center}
		\includegraphics[scale=1.0]{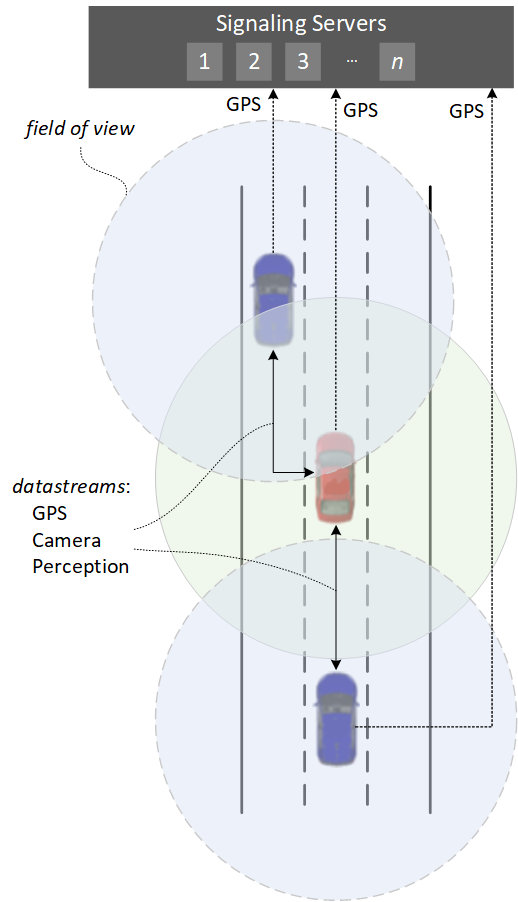}
        \caption{\textbf{Application of CyberCortex.AI to autonomous driving.} Based on GPS data, the signaling servers are automatically interconnecting vehicles located close to each other. Once connected, the vehicles share perception information, thus enabling them to plan their motion based on obstacles which are not visible in their field of view. The red car, located in the center, receives obstacles data from the field of view of the blue cars.}
        \label{fig:use_case_autonomous_driving}
	\end{center}
    \vspace{-8mm}
\end{figure}

To better plan their motion, vehicles equipped with CyberCortex.AI can exchange perceptual and geolocation information. The setup is illustrated in Fig.~\ref{fig:use_case_autonomous_driving}, where the vehicles are sending geolocation information to the Signaling Servers.  When the cars are located within a certain threshold distance to each other, the Signaling Servers will automatically interconnect them. Once connected, the vehicles will exchange GPS, camera and perception information from their respective field of view.

We have chosen the CARLA~\cite{Dosovitskiy17} simulator to replicate the experiments for both competing operating systems. CARLA is an open-source autonomous driving simulator, which enables the simulation of urban layouts, vehicles, pedestrians and buildings, as well as the specification of different sensor suites for the ego car. In the considered examples, we simulate the GPS and the front and rear camera sensors for two cars, while also collecting odometry data. We use the hardware depicted in Fig.~\ref{fig:use_case_surveillance_deployment}(a), as in the previous forest fires prevention use-case.

\begin{figure}
\centering
	\begin{center}
		\includegraphics[scale=0.85]{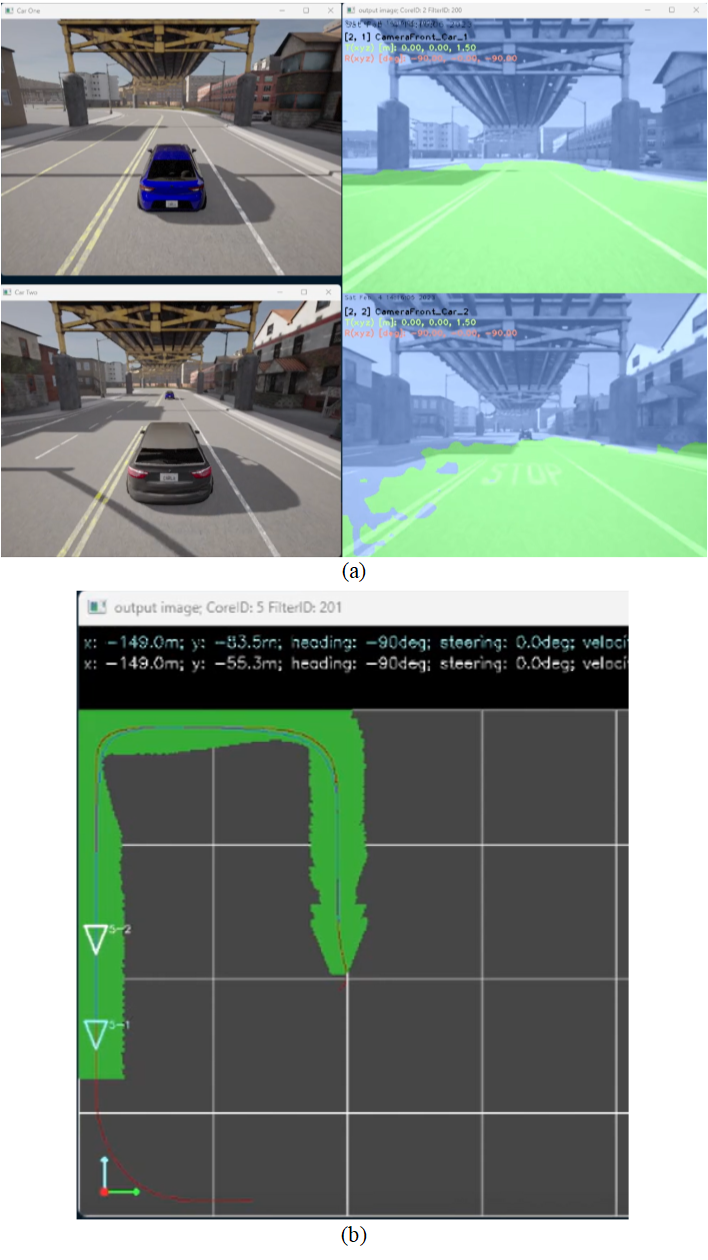}
        \vspace{-4mm}
        \caption{\textbf{Collaborative autonomous driving environment}. (a) Front camera images of the first and second car, respectively, superimposed on the semantic segmentation output. (b) 3D environment model obtained through the reprojection of the semantic segmentation data using the inverse projection matrices of the cameras.}
        \label{fig:carla_perception}
	\end{center}
    \vspace{-4mm}
\end{figure}

The CARLA simulator runs on a desktop PC interfaced with the Nvidia AGX Xavier board and the Raspberry PI computer, each controlling one of the two cars involved in the experiments. The first car is controlled from the Nvidia AGX Xavier board, while the second car is controlled using the Raspberry PI computer. For computational efficiency, both cars use the Nvidia board for DNN inference, as in the previous section.

Four experiments have been performed at different velocities, each over a $1km$ distance. As performance metric, we use the Root Mean Square Error (RMSE) between the trajectory driven using the competing robotics operating systems' control pipelines and a given reference trajectory:

\begin{equation}
	\vspace{-0.2em}
	RMSE = \sqrt{ \sum^{T}_{t=1} \left[ (\hat{p}^{<t>}_x - p^{<t>}_x)^2 + (\hat{p}^{<t>}_y - p^{<t>}_y)^2 \right] },
	\label{eq:rmse}
\end{equation}

\noindent where $\hat{p}^{<t>}_x$, $p^{<t>}_x$, $\hat{p}^{<t>}_y$ and $p^{<t>}_y$ are points on the driven and reference trajectories along the $x-y$ driving plane, respectively.

The drivable area is segmented using a ResNet-50 encoder-decoder neural network~\cite{deeplabv32018}, running on the Nvidia AGX board. Fig.~\ref{fig:carla_perception}(a) shows the camera images, superimposed on the DNN semantic segmentation output, for the two cars located one after another (the blue car visualizes the gray car using its rear camera, while the gray car visualizes the blue car using the front camera). Once segmented, we use the camera's inverse projection matrix to calculate the 3D projection of the drivable area in real-world coordinates. These projections are stored in a 3D octree model of the environment, which is further used for path planning. A snapshot of the 3D model is depicted in Fig.~\ref{fig:carla_perception}(b), where the left area shows the complete model calculated from received images, while the right area show the local environment surrounding the first car.

We use the Dynamic Window Approach (DWA)~\cite{Fox_Dynamic_Window_Approach_1997} to plan and control the motion of the vehicles, based on the segmented drivable area. DWA is an online collision avoidance strategy for mobile robots, which uses robot dynamics and constraints imposed on the robot's velocities and accelerations to calculate a collision free path in the top-view 2D plane. Given an input reference trajectory, the path planner must track the car's motion as close as possible to this reference, while avoiding obstacles.

The operating systems in both vehicles are composed of the following filters, or ROS nodes, respectively:

\begin{itemize}
    \item Front camera filter/node;
    \item Rear camera filter/node;
    \item Depth filter/node;
    \item GPS filter/node;
    \item Perception DNN (ResNet-50 encoder-decoder neural network~\cite{deeplabv32018});
    \item Semantic segmentation filter/node;
    \item DWA path planner filter/node;
    \item Vehicle actuator filter/node.
\end{itemize}

\begin{figure}
	\centering
	\begin{center}
		\includegraphics[scale=0.7]{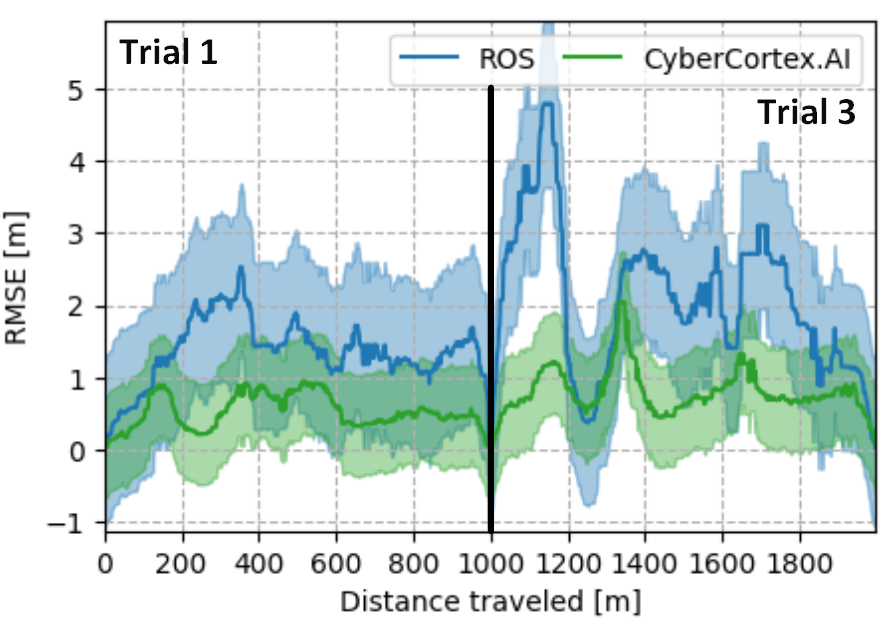}
        \vspace{-4mm}
		\caption{\textbf{RMSE between reference and computed trajectories in trials 1 and 3}. The solid lines indicate the position error, calculated as the RMSE from Eq.~\ref{eq:rmse}, while the shaded regions indicate the standard deviation.}
        \label{fig:carla_position_error}
	\end{center}
    \vspace{-8mm}
\end{figure}

\begin{figure}
	\centering
	\begin{center}
		\includegraphics[scale=0.75]{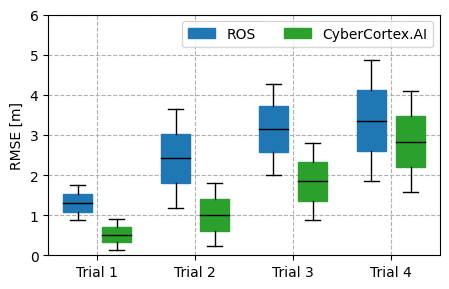}
        \vspace{-8mm}
		\caption{\textbf{Median and variance of RMSE for the four testing scenarios}. Although the RMSE increases proportionally through the four testing trials, it has a lower value for the case of CyberCortex.AI.}
        \label{fig:carla_accuracy_results}
	\end{center}
    \vspace{-8mm}
\end{figure}

The position errors for trials 1 and 3 are shown in Fig.~\ref{fig:carla_position_error}, while Fig.~\ref{fig:carla_accuracy_results} shows the median and standard deviation of the RMSE recorded in the four testing trials.

Fig.~\ref{fig:carla_accuracy_results} indicates that the RMSE is proportional to the vehicles' velocities. However, using CyberCortex.AI we were able to achieve a lower RMSE than in the case of ROS. Since the inner algorithms are the same for both operating systems, we conclude that the lower RMSE obtained using our proposed framework comes from the higher sampling rate of the overall control pipeline. This phenomenon is also visible in Fig.~\ref{fig:carla_position_error}, which shows the RMSE for trials 1 and 3.

\subsection{Discussion}

The two considered experiments have been performed in order to quantitatively and qualitatively evaluate CyberCortex.AI.

Although in our experiments we have used a number of 3 devices, the maximum number of devices allowed when multiple robots are interconnected at the same time is theoretically limited by the internet connection bandwidth and the CPU power. In the Google Chromium implementation of the WebRTC protocol, the number of peers is limited to 256. However, in practical scenarios most peer-to-peer WebRTC applications support up to 20-30 connections. The reason for this limitation is to keep the CPU from overloading from compressing and decompressing video data. If the communcation does not involve video data, as in the case of streaming IMU or state estimation information, then the maximum supported limit of peers should be considerably higher. Additional infrastructure, such as Selective Forwarding Units (SFU) or Multipoint Control Units (MCU), is necesary for supporting a very large number of connections involving heavy video streaming. We consider experimenting with SFUs and MCUs in the future development of CyberCortex.AI.

In order to prevent unwanted devices from linking and reduce bandwidth consumption, we use a mechanism involving a unique communication datastream identifier, where a datastream is uniquely identified by the device's CoreID and the filter's ID.

One of the downsides of the ROS soft synchronization mechanism is that when a node operating at a higher frequency is synchronized with a node operating at a lower frequency, the synchronized output will be sampled at the rate of the low frequency node. Although CyberCortex.AI also provides this kind of soft synchronization, the processing thread within a CyberCortex.AI filter runs at a given sampling rate, which can be set equal to the frequency of the input datastream running at the highest rate. Hence, the filter can process both the high sampled data, as well as the latent incoming information.

Within a DataBlock, the main competition on computing resources takes place between the filters run locally. Resource allocation is currently a limitation of CyberCortex.AI, since the processing power cannot be specified and distributed for each filter. This can be overcome in a future release through the usage of a Real-Time Operating System (RTOS) such as QNX, or a RTOS implementation of embedded Linux.

Another current limitation of CyberCortex.AI is the lack of a functional-safety certification, such as the ASIL A to D  qualifications found in AutoSAR operating systems. Both the RTOS requirements, as well as the ASIL qualifications, are considered in the future development of our software.

\section{Conclusions}
\label{sec:conclusions}

This paper introduces CyberCortex.AI, a lightweight decentralized robotics operating system designed to naturally encapsulate the development, training, deployment and continuous maintenance of artificial intelligence components required within robotics perception-and-control pipelines. We emphasize the importance of data streaming techniques for training deep neural networks, as well as their real-time usage within the AI inference engine. CyberCortex.AI uses the concept of Temporal Addressable Memory (TAM) to manage the high bandwidth and variety of data found in robotics applications.

CyberCortex.AI focuses on the simplicity of designing perception-and-control pipelines, in the form of DataBlocks. We argue that the intrinsic dependencies between the modules are lost if each module is developed independently. This poses negative effects on the control pipeline, in the sense that if one component fails, the overall pipeline will fail. Within CyberCortex.AI, we develop and test processing pipelines as a complete DataBlock, taking into consideration the modules' interdependencies and their sampling rate.




\appendices
\section{Multimedia Description}
\label{appendix:multimedia_description}

A multimedia extension has been prepared to accompany this paper. The extension shows the functionalities of CyberCortex.AI in a comprehensive video, detailing the two use-case scenarios presented in the Performance Evaluation section.

\section{Interactive Demo}
\label{appendix:interactive_demo}

An interactive demonstration has been prepared to accompany this work. It can be accessed through our \url{www.cybercortex.ai} website\footnote{\url{https://cybercortex.ai/login.php}}. Please contact us for access credentials.

The binaries of the CyberCortex.AI.inference system (currently windows only) can be downloaded using the link at the bottom of the page. The demonstrator comes with two example pipelines: \textit{i}) a replay DataBlock of the forest fires prevention system and \textit{ii}) a simulated collaborative autonomous driving system based on CARLA. 

The first example requires the path to the recorded DataBlock, which can be set in the 'drone\_legged.conf' file. The sample dataset can be found at \url{https://tinyurl.com/3wfdevnk}. The replay can be run in command line as:

\vspace{2mm}

App\_CcrCore.exe ..$\backslash$etc$\backslash$pipelines$\backslash$drone\_legged.conf

\vspace{2mm}

The second example requires the RovisLab Carla API bridge\footnote{\url{https://github.com/RovisLab/CarlaAPI}}, used to establish the link between CyberCortex.AI and the CARLA simulated cars. Once the CARLA simulator has been booted from the 'Run\_CarlaClients.py' script, the following 5 DataBlocks can be started:

\vspace{2mm}

App\_CcrCore.exe ..$\backslash$etc$\backslash$pipelines$\backslash$carla\_sim\_car\_1.conf

App\_CcrCore.exe ..$\backslash$etc$\backslash$pipelines$\backslash$carla\_sim\_car\_2.conf

App\_CcrCore.exe ..$\backslash$etc$\backslash$pipelines$\backslash$carla\_ctrl\_car\_1.conf

App\_CcrCore.exe ..$\backslash$etc$\backslash$pipelines$\backslash$carla\_ctrl\_car\_2.conf

App\_CcrCore.exe ..$\backslash$etc$\backslash$pipelines$\backslash$carla\_viz.conf

\vspace{2mm}

The first two DataBlocks are the connections to the simulated cars, used for reading the camera images and states, while providing throttle and steering actuation. The next two DataBlocks are the controllers, used for environment perception and control. The 'carla\_viz.conf' DataBlock is used to aggregate the simulation and control DataBlocks' outputs for visualization.

Once running, the output of each filter can be visualized in command line by typing the number of the filter and pressing carriage return twice. Both pipelines come with a special visualization filter, having ID 200, which synchronizes the outputs of the perception filters. Additionally, the simulated navigation pipeline comes with a filter (ID 201) for visualizing the motion of the vehicle, we well as the tracked keypoints used for simultaneous localization and mapping.

The sampling rate of each filter can be visualized by pressing 't' and carriage return twice. This action will open a window showing the sampling rates as the clock signals from Fig.~\ref{fig:clock_signals_visio}. The total resources consumption, as well as the CPU, GPU and memory consumption of CyberCortex.AI, can be seen the right side of the clock signals.

The DataBlock can also be analyzed using the web debugger. Once a DataBlock has been started, its corresponding filters are registered in the available Signaling Servers, listed on top of the \url{https://cybercortex.ai/droids.php} interface. By clicking on a Signaling Server, one can visualize the datastreams provided by the filters, shown as a list on the website interface.

\bibliographystyle{IEEEtran}
\bibliography{references}

\end{document}